\documentclass[conference]{IEEEtran}
\IEEEoverridecommandlockouts
\usepackage{times}
\usepackage{latexsym}
\usepackage{multirow}
\usepackage{booktabs}
\usepackage{caption}
\usepackage{tabularx}
\usepackage{cite}
\usepackage{amsmath,amssymb,amsfonts}
\usepackage{graphicx}
\usepackage{textcomp}
\usepackage{xcolor}
\usepackage{hyperref}

\def\BibTeX{{\rm B\kern-.05em{\sc i\kern-.025em b}\kern-.08em
    T\kern-.1667em\lower.7ex\hbox{E}\kern-.125emX}}
\begin{document}

\title{Coreference-aware Double-channel Attention Network for Multi-party Dialogue Reading Comprehension}

\author{ Yanling Li\textsuperscript{1},  \ Bowei Zou\textsuperscript{2}, \  Yifan Fan\textsuperscript{1},  \ Mengxing Dong\textsuperscript{1}, \ Yu Hong\textsuperscript{1}$^{\ast}$\thanks{$^{\ast}$Corresponding author.}\\
    \textsuperscript{1}\textit{School of Computer Science and Technology, Soochow University}, Soochow, China \\
    \textsuperscript{2}\textit{Institute for Infocomm Research, A*STAR}, Singapore \\
    \texttt{\{li4861988, yifanfannlp, ayumudong, tianxianer\}@gmail.com} \\
    \texttt{zou\_bowei@i2r.a-star.edu.sg}
  }

\maketitle

\begin{abstract}
 We tackle Multi-party Dialogue Reading Comprehension (abbr., MDRC). MDRC stands for an extractive reading comprehension task grounded on a batch of dialogues among multiple interlocutors. It is challenging due to the requirement of understanding cross-utterance contexts and relationships in a multi-turn multi-party conversation. Previous studies have made great efforts on the utterance profiling of a single interlocutor and graph-based interaction modeling. The corresponding solutions contribute to the answer-oriented reasoning on a series of well-organized and thread-aware conversational contexts. However, the current MDRC models still suffer from two bottlenecks. On the one hand, a pronoun like ``{\em it}'' most probably produces multi-skip reasoning throughout the utterances of different interlocutors. On the other hand, an MDRC encoder is potentially puzzled by fuzzy features, i.e., the mixture of inner linguistic features in utterances and external interactive features among utterances. To overcome the bottlenecks, we propose a coreference-aware attention modeling method to strengthen the reasoning ability. In addition, we construct a two-channel encoding network. It separately encodes utterance profiles and interactive relationships, so as to relieve the confusion among heterogeneous features. We experiment on the benchmark corpora Molweni and FriendsQA. Experimental results demonstrate that our approach yields substantial improvements on both corpora, compared to the fine-tuned BERT and ELECTRA baselines. The maximum performance gain is about 2.5\% $F$1-score. Besides, our MDRC models outperform the state-of-the-art in most cases.

\end{abstract}


\renewcommand{\thefootnote}{}
\footnotetext{The research is supported by National Key R\&D Program of China (2020YFB1313601) and National Science Foundation of China (62076174, 61836007).}

\section{Introduction}

\begin{figure*}[t]
    \centering
    \includegraphics[width=0.80\textwidth]{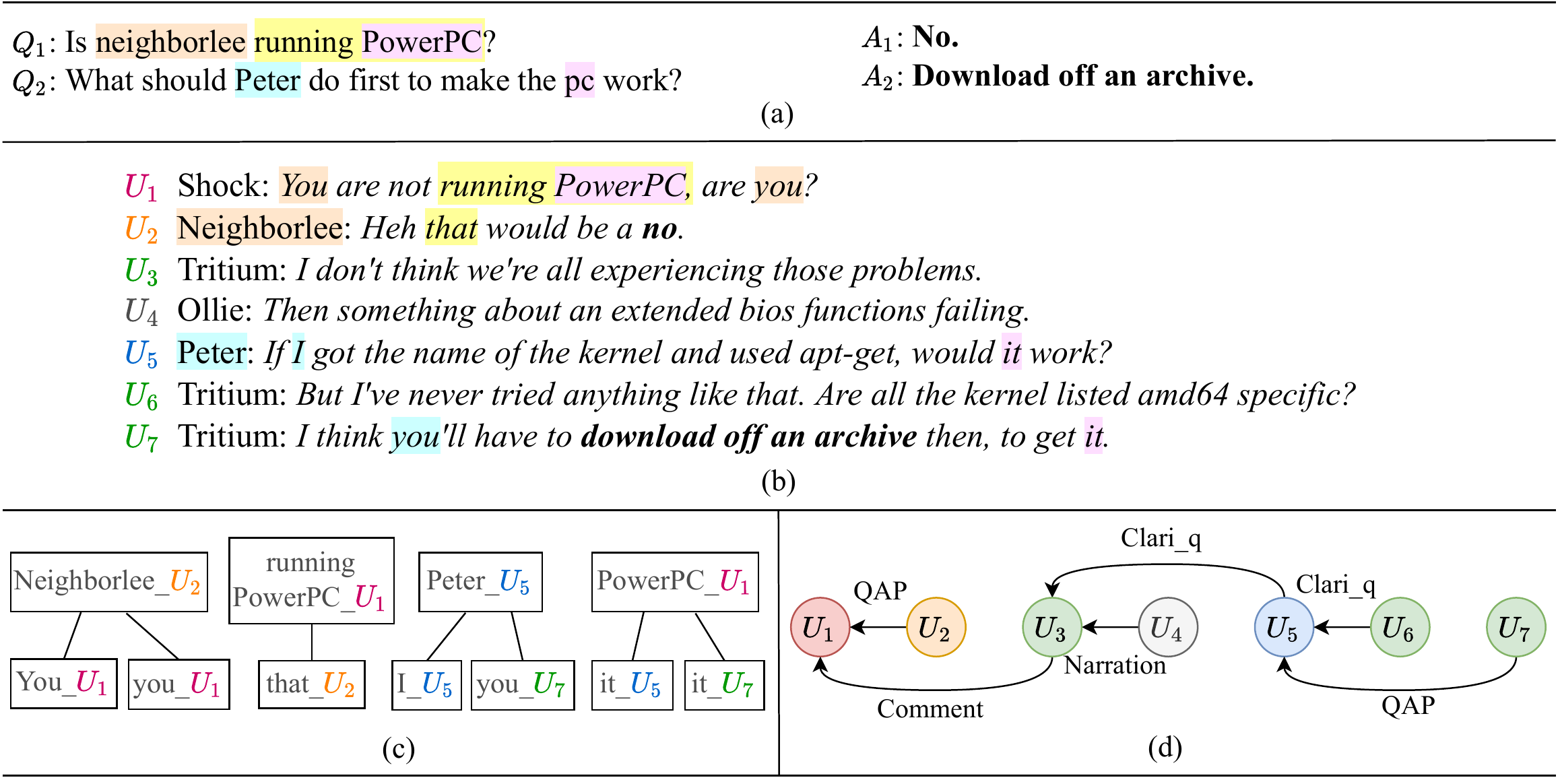}
    \caption{\small{(a) Two question-answering pairs of the dialogue in (b). (b) An multi-party dialogue from Molweni \cite{li2020molweni}. Different highlight colors indicate different coreference clusters required for (a). (c) Different coreference clusters of the example dialogue. (d) The discourse graph of the example dialogue, where utterances with the same interlocutor are filled with the same color.}}
    \label{figure1}
    \vspace{-2mm}
\end{figure*}

MDRC \cite{sun2019dream,cui2020mutual} is an increasingly attractive task in the field of multi-party dialogue research \cite{sun2019dream,ma2018challenging,yang2019friendsqa,li2020molweni}. Essentially, it serves as an extractive Machine Reading Comprehension task (MRC), automatically extracting proper answers from the given context for a series of questions. Though, it is more challenging than the conventional MRC because the context consists of multi-party dialogues, instead of a narrative or two-party conversations. The information flow that an MDRC model needs to treat is generally characterized by irregular and unstructured utterances, i.e., all sorts of gossip among different interlocutors (see the examples in Fig.~\ref{figure1}-({\tt a}) and ({\tt b})).
% ({\tt a}) and ({\tt b})). 

Similar to nearly the majority of current MRC solutions, the studies of MDRC leverage encoder-decoder neural networks. The encoder is generally implemented using Pre-trained Language Models (PLMs) \cite{kenton2019bert,liu2019roberta,lan2019albert,clark2020electra,joshi2020spanbert}, which learns to represent the sequentially-occurred utterances with attention mechanisms \cite{cheng2016long,vaswani2017attention}. The decoder is developed to conduct token-level boundary discrimination, i.e., positioning the boundaries of answers grounded on the attentive representations of utterances.

Recently, researchers tend to give deeper insights into the unique characteristics of MDRC and, accordingly, concentrate on two task-specific schemes, including utterance profiling and interaction modeling:

\begin{itemize}
    \item {\textbf {Utterance profiling}} is designed to collect the utterances of a single interlocutor, and encode them separately. The goal is to represent distinguishable internal features (e.g., coherence) in the dialogue history of the same interlocutor. It helps to bridge discrete clues for answer prediction, with less distraction from the utterances of other interlocutors.  
    \item {\textbf {Interaction modeling}} is conditioned on the analysis of discourse dependency relationships among utterances of all interlocutors, instead of that of a single interlocutor. Interactively modeling this kind of global structural features (e.g., posting and response) enables the connection of clues hidden in different interlocutors' utterances.
\end{itemize}

Both of the aforementioned approaches contribute to the enhancement of the existing MDRC models (as overviewed in Section $\uppercase\expandafter{\romannumeral2}$). However, they still suffer from two bottlenecks. First, some of the crucial clues for reasoning are actually co-referred across utterances, or maintaining close relationships with pronouns. For example, the highlighted clues like ``{\em running PowerPC}'' in Fig.~\ref{figure1}-({\tt c}) is co-referred by the pronoun ``{\em that}'' in the utterance $U_2$, while the answer ``{\em No}'' in this case syntactically depends on the pronoun heavily (see $Q_1$, $U_1$ and $U_2$ in Fig.~\ref{figure1}-({\tt a}) and ({\tt b})). Though, the co-reference resolution has neither been incorporated into the utterance profiling process nor interaction modeling. Second, most of the current studies fail to simultaneously take advantage of utterance profiling and interaction modeling. Instead, they are conducted separately and independently. 
This unavoidably causes the omission of dependency features of either internal clues or external ones. 

To overcome the bottlenecks, we propose a Coreference-Aware Double-channel Attention network (abbr., CADA). CADA is first characterized by the awareness of coreference at the initial encoding stage. Coreference resolution is performed to formulate the one-hot coreference graph. Graph-based attention modeling is further used to enhance token-level semantic encoding over all the utterances, where PLMs are considered as the fundamental encoders. 

CADA is additionally equipped with two parallel encoding channels. One performs utterance profiling to model attentive dependency features of internal clues within the utterances of each interlocutor, where a role-based utterance coherence graph is utilized. The other channel conducts interaction modeling to strengthen the interaction representation among multi-party utterances. Essentially, it is applied to perceive the attentive dependency features of external clues. Discourse-level dialogue graph is used for interaction modeling. We finally combine the three encoding results for answer prediction.

In our experiments, we construct two CADA models using BERT \cite{kenton2019bert} and ELECTRA \cite{clark2020electra} respectively. We evaluate models on the benchmark corpora Molweni \cite{li2020molweni} and FriendsQA \cite{yang2019friendsqa}, in terms of the canonical data partition schemes. Experimental results show that our approach yields significant improvements compared to the BERT and ELECTRA baselines. The maximum performance gain is about 2.5\% $F$1-score. In particular, CADA outperforms the state-of-the-art MDRC models.

The rest of the paper is organized as follows. Section $\uppercase\expandafter{\romannumeral2}$ overviews the related work. We present CADA in Section $\uppercase\expandafter{\romannumeral3}$. Section $\uppercase\expandafter{\romannumeral4}$ gives the experimental results and corresponding discussions on them. We conclude the paper in Section $\uppercase\expandafter{\romannumeral5}$.

\section{Related Work}
We briefly overview two categories of related work, including that of dialogue machine reading comprehension and dialogue discourse dependency. The study of dialogue discourse analysis provides fundamental theories and supportive techniques for modeling cross-utterance interaction.
\subsection{Dialogue Machine Reading Comprehension}
Recently, the research of dialogue-oriented linguistic computing is undergoing a new trend, i.e., performing MRC over complicated multi-party dialogues \cite{sun2019dream,yang2019friendsqa,li2020molweni}. This raises a new research topic---MDRC. The most crucial challenge of MDRC is that the interactions among different interlocutors produce an intricate discourse information flow, and thus the current semantics encoding approaches suffer from the confusion of internal features (i.e., the ones occurring in the utterances of a single interlocutor) and external features (i.e., the ones caused by interaction with other interlocutors).

Liu et al. \cite{liu2021filling} propose a decoupled fusion network to decouple interlocutor information from utterance receiver and sender, where masked attention mechanisms are utilized. Ma et al. \cite{ma2021enhanced} strengthen the aforementioned mechanism using a speaker-aware graph. The graph explicitly represents the consanguineous relations for the utterances of the same speaker. Li et al. \cite{li2021self} design a self-supervised task to model interlocutors' information flows implicitly. All the above studies prove the awareness of interlocutors' characteristics and internal features of their own utterances contributes to dialogue understanding. More importantly, Ma et al. \cite{ma2021enhanced} and Li et al. \cite{li2021self} successfully use utterance profiling to improve the current MDRC models.

It is noteworthy that our approach is different from the previous work. We enhance utterance profiling conditioned on the interlocutor role information. More importantly, we combine internal features which are pronouns and their corresponding entities, where coreference resolution is used.

\begin{figure*}[t]
    \centering
    \includegraphics[width=0.78\textwidth]{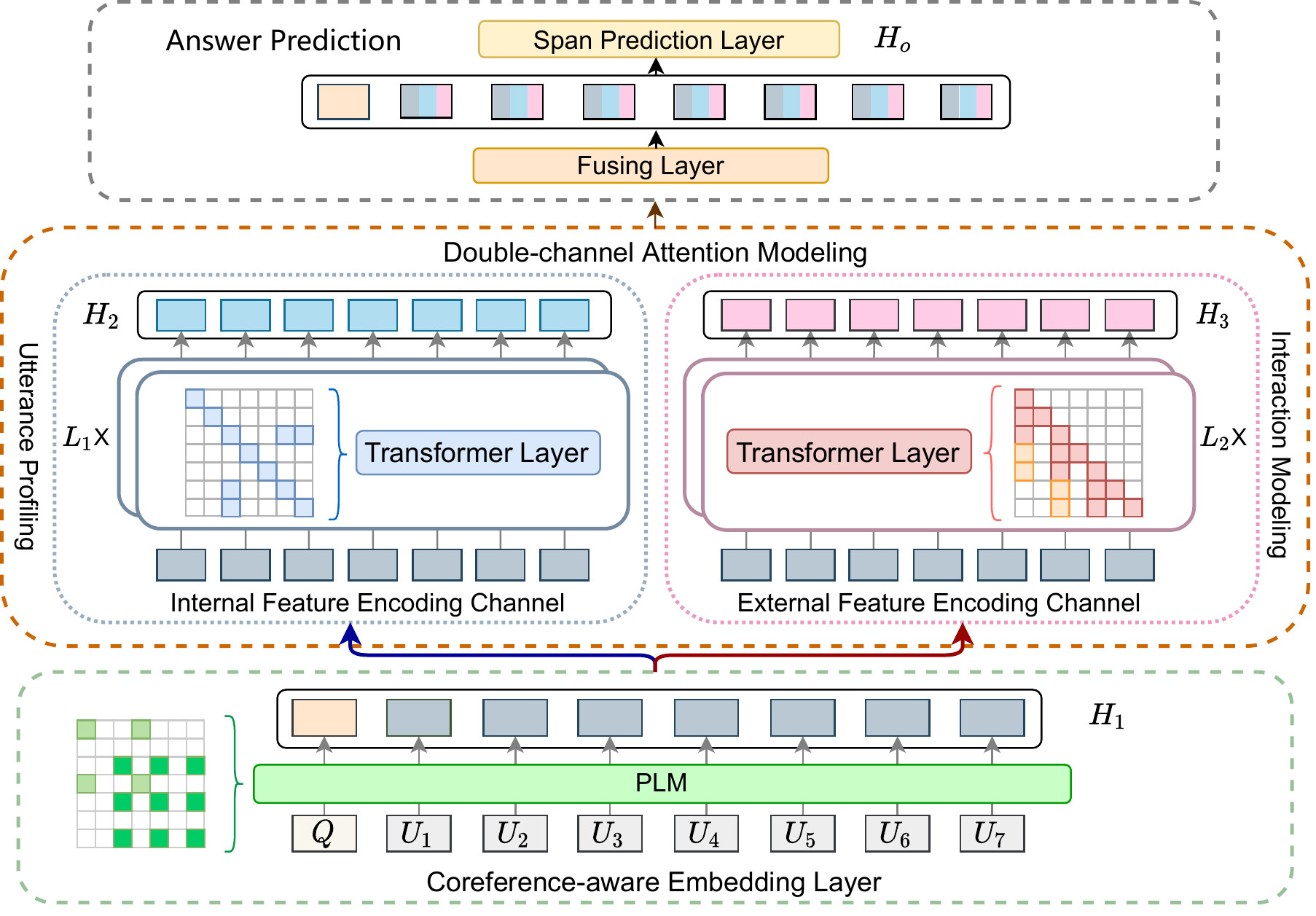}
    % \vspace{-6mm}
    \caption{\small{The main architecture of Coreference-Aware Double-channel Attention network (CADA).}}
    \label{figure2}
    \vspace{-2mm}
\end{figure*}

\subsection{Dialogue Discourse Dependency}
In general, an interlocutor may respond to various utterances of other interlocutors in a multi-party dialogue flow. This allows the utterance-hopping discourse dependency structure to be formed implicitly. The previous studies illustrate that discourse dependency parsing helps to perceive the interaction mode among interlocutors and, accordingly, the potentially connected clues for reasoning \cite{asher2016discourse,gao2020discern,jia2020multi,chen-yang-2021-structure,ma-etal-2022-structural}.

Recently, a variety of utterance-level discourse parsers have been proposed. Shi and Huang \cite{shi2019deep} propose a deep sequential model for dependency parsing over utterances. Liu et al. \cite{liu2021improving} design a transformer-based discourse parser. It improves cross-domain performance using prior language knowledge and cross-domain pre-training. These parsers provide explicit discourse structures of utterances for downstream applications.

In addition, considerable efforts have been made to integrate discourse structure into dialogue understanding models. Li et al. \cite{li2021dadgraph} develop a graph-based model (namely DADgraph) to fuse discourse dependency features, where a graph neural network is used. Jia et al. \cite{jia2020multi} design a thread encoder that incorporates dialogue dependency information by threads. On this basis, PLM is leveraged to generate corresponding representations. Gu et al. \cite{gu2021mpc} develop an MPC-BERT model trained in several self-supervised tasks, where discourse structural features are used as the objectives for self-supervised learning. 
% The considered tasks comprise reply-to-utterance recognition, pointer consistency distinction, and so on.

In this paper, we incorporate discourse structure information into an attention-based encoding channel of CADA. The goal is to enhance the capacity of CADA in perceiving the close interaction and, accordingly, assign higher attention to the interactively-dependent constituents among utterances.

\section{Approach}
We develop a coreference-aware two-channel neural network (namely CADA) for MDRC. CADA encodes the input utterances in terms of the raised question as usual. Though, it performs encoding to represent two categories of features, including internal and external features. The internal features are taken by utterance profiling in the coreference-aware embedding layer and a separate encoding channel, where coreference resolution and interlocutor role information are used to enhance the attention modeling of token-level semantic relationships, respectively. 
The external features are extracted by interaction modeling in another separate encoding channel, where explicit discourse dependency structures are used for modeling attentive information that is bridged by strong interactive relationships (among different interlocutors). Structurally, the two-channel neural encoders are built over a PLM-based embedding layer. PLMs like BERT and ELECTRA are used in our experiments. 
Besides, the encoders are connected with a fusion layer and the Fully-Connected linear layer (FC layer). The FC layer plays the role of a decoder which predicts the answers. We show the architecture of CADA in Fig.~\ref{figure2}.

\subsection{Input and Output of MDRC}
We follow the previous studies~\cite{yang2019friendsqa,li2020molweni} to set up the input and output formations. Given a multi-party dialogue context $D = \{U_1, U_2, ..., U_n\}$ with $n$ utterances and a question $Q$, MDRC aims to extract a text span $\mathcal{A}$ from $D$, and specify it as the answer for the question $Q$. Each utterance $U_i=\{S_i,W_i\}$ includes a interlocutor name $S_i$ and utterance contents $W_i$ issued by $S_i$. We denote $W_i$ as a word sequence $ W_i = \{w_{i1}, w_{i2}, ..., w_{il}\}$ with $l$ tokens.

It is noteworthy that, in some benchmark MDRC corpus, the answer $\mathcal{A}$ can be answerable or unanswerable. The answerable $\mathcal{A}$ is constituted with a concrete text span (one token at least) and the unanswerable $\mathcal{A}$ has been officially designated with a tag ``{\em Truly Impossible}''.

\subsection{Coreference-aware Embedding Layer}
We conduct the first-stage encoding at the embedding layer, where PLM is utilized for computing the initial distributed representations of tokens in both $Q$ and $D$. Different from the previous work, we additionally use coreference resolution at this stage, where pronouns correspond to the referred entities (nouns in some cases). This enables the sharing of attentive information among the co-referred tokens. 

\begin{figure}[t]
    \centering
    \includegraphics[width=0.5\textwidth]{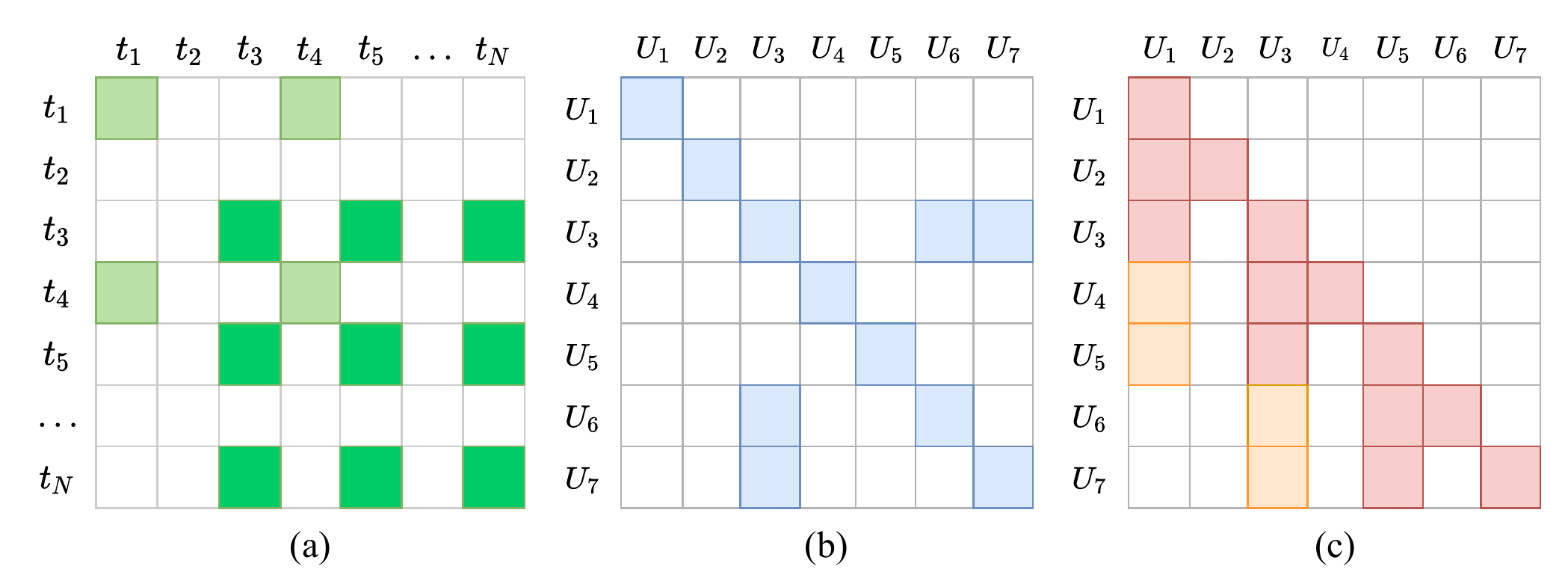}
    \caption{\small{(a) An example of coreference adjacent matrix $M_{1}$, where different depths of green color indicate different coreference clusters. For instance, $\{t_1, t_4\}$ belong to one cluster and $\{t_3, t_5, t_N\}$ relate to another, while $t_2$ has no coreference link. (b) and (c) show the role-based utterance matrix $M_2$ and discourse dependency matrix $G$ for dialogue in Fig.~\ref{figure1}, respectively. For the sake of simplicity, we display in form of utterance units in (b).
    Different colors in (c) denote different path lengths between utterances in the discourse graph. We set the threshold $\gamma$ as 2 in (c) for convenience.}}
    % \vspace{-3mm}
    \label{figure3}
\end{figure}

Specifically, we first concatenate $Q$ and utterances in $D$ to form the input sequence $X$ = \{[CLS] $Q$ [SEP] $U_1$ [SEP] $U_2$ \,...\,[SEP] $U_n$ [SEP]\}, where [CLS] and [SEP] are special tokens. The token [SEP] is used to separate adjacent utterances. Further, we employ the coreference resolution toolkit \cite{joshi2020spanbert,lee2018higher} to cluster pronouns and entities they referred (namely coreference cluster for short). For each dialogue $D$, in general, there are a variety of coreference clusters $\{C_1, C_2,..., C_u\}$ that can be discovered and taken out of the utterances. For each cluster $C_i$, the mentions (pronouns) referring to the same entity are organized into the sequence $e_i=\{e_{i1}, e_{i2},...,e_{im}\}$, where $e_{ij}$ is a mention or the referred entity. We use the graph to represent the coreference relationships of all the sequences $e_i$s, i.e., one subgraph per $e_i$ and one global graph obtained by concatenating all subgraphs. Mathematically, the one-hot adjacent matrix $M_{1} \in R_{N\times N}$ is used to represent the global graph of the input sequence $X$, where $N$ denotes the number of tokens in $X$.
We show an example of $M_{1}$ in Fig.~\ref{figure3}-({\tt a}), where two subgraphs exist. Each corresponds to a coreference cluster. The coreference relations are highlighted by green colors which are signaled by the value ``1'' mathematically in the adjacent matrix (Note the coreference relations in the subgraphs are highlighted with different depths of green color).

To enhance token-level coreference-aware encoding, we use $M_{1}$ to incorporate additional graph-based attention into the encoding process.
Inspired by the previous study \cite{xu2021entity}, we set up the trainable graph-based coreference-aware attention mechanism, and deploy it together with the canonical multi-head self-attention mechanism. The goal is to enable the direct learning of structural coreference information conditioned on the query and key representations. Specifically, we utilize biaffine transformation \cite{yu2020named} to construct coreference attentive score $\lambda$. It is computed as \eqref{eq1}, where $W$ is a trainable matrix, and $b$ serves as a prior bias which is utilized for differentiating coreference dependency from other features in the dialogue context. Accordingly, the coreference-aware attention score between the $i$-th and $j$-th token in each attention head is calculated as \eqref{eq2}.
\begin{equation}\label{eq1}
\begin{split}
    \lambda_{ij}=q_iW_{ij}k_j^{T}+b_{ij}, 
\end{split}
\end{equation}
\begin{equation}\label{eq2}
\begin{split}
    e_{ij}= \frac{q_ik_{j}^{T}+\lambda_{ij}M_{1,ij}}{\sqrt{d}},
\end{split}
\end{equation}
By the aforementioned computation, we finally obtain the coreference-aware distributed representation $H_1\in R_{N\times F}$, where $F$ denotes hidden states dimension.

\subsection{Double-channel Attention Modeling}
Grounded on the coreference-aware embedding layer, we develop a double-channel attention-based encoding model for utterance profiling and interaction modeling. Specifically, it is implemented using an interlocutor property module and discourse dependency module, respectively.

\subsubsection{Interlocutor Property Module}
Due to the uncertainty of the interaction process, there are intricate interlocutor information flows in a multi-party conversation. For example, five interlocutors participate in the dialogue in a relatively random order in Fig.~\ref{figure1}. Meanwhile, the specific interlocutor information of ``{\em neighborlee}'' and ``{\em Peter}'' in the case is required during answering $Q_1$ and $Q_2$, respectively.
However, the information actually exchanges among interlocutors. As a result, the essential information flows back and forth across interlocutors and utterances. This phenomenon makes it difficult to either reason about key clues or capture the specific property of each interlocutor.
Therefore, we set up an independent encoding channel to model the interlocutor property information. It conducts utterance profiling to represent the internal property of each interlocutor in the dialogue.

All the utterances issued by the same interlocutor imply individual (personal) features, and they naturally form a role-based utterance coherence subgraph. Accordingly, we build a role-based utterance matrix $M_{2} \in R_{N\times N}$, which is illustrated in Fig.~\ref{figure3}-({\tt b}). More specifically, we denote ${s}_k$ as the interlocutor of the $k$-th utterance, and ${k}_i$ as the utterance where the $i$-th token is located. We set $M_{2}[i,j]=1$ if $s_{k_i}=s_{k_j}$. Hence, $M_{2}$ is formulated as follows: 
\begin{equation}\label{eq3}
    \begin{aligned}
            M_{2}[i,j] & = 
            \left\{
            \begin{array}{rcl}
            1,       &      & {s_{k_i}=s_{k_j}}\\
            0,     &      & {otherwise}
            \end{array} 
            \right.
    \end{aligned}
\end{equation}

The graph-based attention is used to regulate the interaction with the interlocutor property information. In detail, we employ biaffine transformation \cite{yu2020named} and the aforementioned $M_{2}$ to construct interlocutor property attention $\dot{e}_{ij}$. The calculation method of $\dot{e}_{ij}$ is similar to \eqref{eq1} and \eqref{eq2}. The only change is that $M_{1,ij}$ is necessarily substituted by $M_{2,ij}$. Further, we use $L_1$-layer transformer  ${f}_{tran}$ to compute the representations $H_2\in R_{N\times F}$ of interlocutor properties: $H_2$= ${f}_{tran}(H_1, \dot{e})$.

\subsubsection{Discourse Dependency Module}
As discussed in Section $\uppercase\expandafter{\romannumeral2}(B)$, discourse dependency structure exhibits the interactive dependencies among utterances of different interlocutors. Fig.~\ref{figure1}-({\tt d}) shows an example of the discourse structure graph, where each utterance serves as a node, and each node is connected by at least one specific-type relation arc. Such structure is crucial for perceiving valuable clues. For example, the fifth and seventh utterances (i.e., $U_5$ and $U_7$) hold a QAP relation (i.e., Question-Answering-Pair relation), which potentially contributes to the perception of clues for answering $Q_2$. Accordingly, we intend to use the discourse dependency structure of dialogue to attentively weigh the utterances that possess key clues. Correspondingly, we design a decoupling module and deploy it at the second encoding channel, so as to impose the effects of discourse structure information upon the token-level representations. 

To be specific, we follow Ying et al. \cite{ying2021transformers} to compute the distance between utterances in the dependency graph, and use it to establish a discourse dependency matrix $G$. For example, in Fig.~\ref{figure1}-({\tt b}), $U_7 \to U_5 \to U_3$ is a path being launched from $U_7$ to $U_3$, which holds a length $l$ of 2 ($l_{37}$=2) in terms of the relation arcs ``QAP'' and ``Clari\_q (i.e., Clarification question)''. Besides, we set $l_{ij}$ to 1 for the path between an utterance and itself. It is noteworthy that we set $G_{ij}$ to 0 when building $G$ if the length $l_{ij}$ is larger than 0, while we set $G_{ij}$ to $-\infty$ when there is no path between utterances (i.e., $G_{ij}$=$-\infty$ if $l_{ij}$=0). For the utterances that are far apart from each other, we assume that their dependency intensity is relatively weak and negligible. Accordingly, we also set $G_{ij}$ to $-\infty$ if the length between utterances is larger than the threshold $\gamma$.  Fig.~\ref{figure3}-({\tt c}) shows an example for matrix $G$. We formulate the dependency matrix $G$ as follows:
\begin{equation}\label{eq4}
    \begin{aligned}
            G[i,j] & = 
            \left\{
            \begin{array}{rcl}
            0,   &   &  0 < {l_{i,j} \le \gamma}\\
            {-\infty},     &      & {otherwise}
            \end{array} 
            \right.
    \end{aligned}
\end{equation}
 
Note that the connectivity itself (i.e., whether dependent) is more crucial than the dependency type (i.e., marks on dependency arcs) for our method. Hence, we omit the representation of dependency types when building dependency graphs.

We utilize $L_2$-layer transformer blocks to incorporate discourse dependency information into the interaction encoding channel. Following the scheme of masking attention \cite{liu2021filling}, we adopt a mask-based multi-head self-attention mechanism to emphasize correlations between utterances:
\begin{equation}\label{eq5}
\begin{split}
    &Atten\left ( Q,K,V \right )=softmax\left (\frac{QK^{T}}{\sqrt{d_{k}}}+G\right )V,
\end{split}
\end{equation}
where $Q, K, V$ denote the query, key, and value representation, respectively, and $d_k$ is the dimension of the key. Eventually, we obtain the distributed representation $H_3\in R_{N\times F}$ that involves the attention-weighted effects of discourse dependency information.

\subsection{Answer Prediction and Training}
We combine the aforementioned three representations (i.e., coreference-aware representation $H_1$, as well as the updated representations $H_2$ and $H_3$), and feed the fused representation to the decoder for predicting answers. Specifically, we concatenate $H_1, H_2, H_3$ to form the final representation $H_o\in R_{N\times 3F}$:
\begin{equation}\label{eq8}
    H_o = [H_1, H_2, H_3],
\end{equation}

On this basis, we use two fully-connected linear layers to compute the probability $p^{s}$ of the start position of a possible answer span over all tokens in the dialogue, as well as the probability $p^{e}$ of the end position. Answer extraction is conducted by taking out tokens between start and end positions. Besides, we employ a linear classifier to estimate the binary probability $p^{t}$ of whether the question is answerable. Given the start and end position of the answer span $[\mathcal{A}_{s}, \mathcal{A}_{e}]$ and answer type $\mathcal{A}_{t}$, the computation is formulated as follows:
% we use cross-entropy loss to train CADA:
\begin{equation}\label{eq9}
\begin{split}
    p^{s} = softmax\left ( H_oW_{s}+b_{s} \right ), \\
    p^{e} = softmax\left ( H_oW_{e}+b_{e} \right ), \\
    p^{t} = sigmoid\left ( H_oW_{t}+b_{t} \right ), \
\end{split}
\end{equation}
where $W_s, W_e, W_t, b_s, b_e, b_t$ are trainable matrices and biases. 

We use cross-entropy loss of both answer type and answer itself to train CADA. The loss is computed as \eqref{eq12}, where $K$ denotes the number of examples in a batch.
\begin{equation}\label{eq12}
    Loss = -\frac{1}{K}\sum_{K}^{} [log(p^s_{\mathcal{A}_{s}}) + log(p^e_{\mathcal{A}_{e}}) + log(p^t_{\mathcal{A}_{t}})],
\end{equation}

\section{Experimentation}
We report the test results in this section, along with the experimental settings including the benchmark corpora, evaluation metrics, and implementation details (i.e., hyperparameter settings). We also conduct a series of analyses and discussions over the experimental results in this section.

\subsection{Datasets}
We experiment on two benchmark MDRC corpora, including Molweni \cite{li2020molweni} and FriendsQA \cite{yang2019friendsqa}. Molweni is the first MDRC dataset that annotates the discourse dependency structure of dialogues, which is derived from Ubuntu Chat Corpus \cite{lowe2015ubuntu}. The interlocutor number per dialogue in Molweni is 3.51 on average. Besides, Molweni includes both answerable extractive questions and unanswerable cases. Due to the exact annotation of discourse dependency structure and the real scenario of multiple-interlocutor dialogue, Molweni appears as an appropriate corpus for the research of MDRC that is conditioned on the ground-truth structural information. FriendsQA is a question-answering dataset excerpted from the TV show \textit{Friends}. Most of the dialogues in it are colloquial everyday conversations. FriendsQA contains 1,222 dialogues and 10,610 answerable extractive questions. Different from Molweni, FriendsQA does not provide the ground-truth annotations of discourse dependency structure among utterances. Therefore, We use Liu et al. \cite{liu2021improving}'s parser to pretreat the dialogues in FriendsQA for acquiring the structural information.

We follow the canonical partition schemes to divide both corpora into the training, validation (Dev), and test sets. The statistics of the datasets are shown in Table \ref{tab0}.

\begin{table}[t]
\centering
\caption{\small{Statistics in Molweni and FriendsQA.}}\label{tab0}
\tabcolsep=0.1cm
\begin{tabular}{c|ccc|ccc}
\toprule[1pt]
\multirow{2}{*}{Datasets} & \multicolumn{3}{c|}{Molweni}                 & \multicolumn{3}{c}{FriendsQA}                                               \\ & \multicolumn{1}{c}{Train} & \multicolumn{1}{c}{Dev} & Test & \multicolumn{1}{c}{Train} & \multicolumn{1}{c}{Dev} & Test \\ \hline 
Dialogues                  & \multicolumn{1}{c}{8,771}     & \multicolumn{1}{c}{883}     & 100    & \multicolumn{1}{c}{977}       & \multicolumn{1}{c}{122}     & 123     \\ 
Utterances               & \multicolumn{1}{c}{77,374}       & \multicolumn{1}{c}{7,823}      & 2,513     & \multicolumn{1}{c}{21,607}       & \multicolumn{1}{c}{2,847}      & 2,336     \\ 
Questions                      & \multicolumn{1}{c}{24,682}       & \multicolumn{1}{c}{2,513}      & 2,871     & \multicolumn{1}{c}{8,535}       & \multicolumn{1}{c}{1,010}      & 1,065     \\
\bottomrule[1pt]
\end{tabular}
\end{table}

\subsection{Comparison and Evaluation}
We use two PLMs as the baselines, including BERT$_{large}$ \cite{kenton2019bert} and ELECTRA$_{large}$ \cite{clark2020electra}. Both are connected with multiple perceptions for decoding the answers. In addition, we compare our CADA to two state-of-the-art neural MDRC models, including SKIDB \cite{li2021self} and ESA \cite{ma2021enhanced}. We follow the common practice \cite{li2020molweni,li2021self} to use $F$1-score and exact matching score (EM for short) to evaluate all the models in our experiments.

\begin{table}[t]
\centering
% \small 
    \caption{\small{The test results on Molweni and FriendsQA. The mark ``$\dagger$'' denotes a statistically significant improvement in the $F$1-score (p $<$ 0.05) compared with baselines.}}\label{tab1}
\tabcolsep=0.1cm
\begin{tabular}{lllll}
\toprule[1pt]
\multirow{2}{*}{Model}   & \multicolumn{2}{c}{Molweni} & \multicolumn{2}{c}{FriendsQA} \\ & \multicolumn{1}{c}{EM}  & \multicolumn{1}{c}{F1}  &  \multicolumn{1}{c}{EM}   & \multicolumn{1}{c}{F1} \\ \hline
\multicolumn{5}{c}{$BERT$ (Devlin et al., 2019) \cite{kenton2019bert}}    \\
\multicolumn{1}{l}{BERT (Baseline)}   & 
\multicolumn{1}{c}{50.5}    & \multicolumn{1}{c}{65.1}          & \multicolumn{1}{c}{46.0}  & \multicolumn{1}{c}{63.1}     \\ 
\multicolumn{1}{l}{SKIDB (Li and Zhao, 2021) \cite{li2021self}}   &
\multicolumn{1}{c}{51.1}          & \multicolumn{1}{c}{66.0} & \multicolumn{1}{c}{46.9}          & \multicolumn{1}{c}{63.9} \\ 
\multicolumn{1}{l}{ESA (Ma et al., 2021) \cite{ma2021enhanced}}  &
\multicolumn{1}{c}{52.9}            & \multicolumn{1}{c}{66.9}  & \multicolumn{1}{c}{\textbf{49.0}} & \multicolumn{1}{c}{64.0}  \\ 
\multicolumn{1}{l}{${\rm {CADA}}^\dagger$ (Ours)}        &
\multicolumn{1}{c}{\textbf{52.9}}  & \multicolumn{1}{c}{\textbf{67.6}} &  \multicolumn{1}{c}{47.4}   & \multicolumn{1}{c}{\textbf{65.6}}   \\ \cline{1-5}
\multicolumn{5}{c}{$ELECTRA$ (Clark et al., 2020) \cite{clark2020electra}}          \\ 
\multicolumn{1}{l}{ELECTRA (Baseline)}    &
\multicolumn{1}{c}{57.4}          & \multicolumn{1}{c}{71.8}       & \multicolumn{1}{c}{56.9}          & \multicolumn{1}{c}{74.9}  \\ 
\multicolumn{1}{l}{SKIDB (Li and Zhao, 2021) \cite{li2021self}}   &
\multicolumn{1}{c}{58.0}          & \multicolumn{1}{c}{72.9}       & \multicolumn{1}{c}{55.8}          & \multicolumn{1}{c}{72.3}     \\ 
\multicolumn{1}{l}{ESA (Ma et al., 2021) \cite{ma2021enhanced}} &
\multicolumn{1}{c}{58.6}          & \multicolumn{1}{c}{72.2}        & \multicolumn{1}{c}{58.7} & \multicolumn{1}{c}{75.4}    \\ 
\multicolumn{1}{l}{${\rm {CADA}}^\dagger$ (Ours)}     &
\multicolumn{1}{c}{\textbf{59.8}}   &\multicolumn{1}{c}{\textbf{73.6}}& \multicolumn{1}{c}{\textbf{59.2}}          & \multicolumn{1}{c}{\textbf{76.7}}        \\ \bottomrule[1pt]
\end{tabular}
% \vspace{-2mm}
\vspace{-2mm}
\end{table}

\begin{table}[t]
\centering
\caption{\small{Performance of ELECTRA-based CADA on different question types. The marks $\uparrow$ and $\downarrow$ denote performance gain and induction compared to the ELECTRA baseline. Dist signals the proportion of each question type in the dataset (\%).}}\label{tab22}
\tabcolsep=0.1cm
\begin{tabular}{l|lll|lll}
\toprule[1pt]
\multicolumn{1}{c|}{\multirow{2}{*}{Types}}  & \multicolumn{3}{c|}{Molweni}    & \multicolumn{3}{c}{FriendsQA}   \\ &
 \multicolumn{1}{c}{Dist.} & \multicolumn{1}{c}{EM}         & \multicolumn{1}{c|}{F1}      & \multicolumn{1}{c}{Dist.} & \multicolumn{1}{c}{EM}          & \multicolumn{1}{c}{F1}    \\ \hline
\multicolumn{1}{c|}{Who}   & \multicolumn{1}{c}{4.7}   & \multicolumn{1}{c}{80.1~($\uparrow 4.2$)} & 82.1~($\uparrow 4.8$)  & \multicolumn{1}{c}{18.8} & \multicolumn{1}{c}{66.4~($\uparrow 2.2$)}  & 78.4~($\uparrow 1.9$) \\ 
\multicolumn{1}{c|}{When}  & \multicolumn{1}{c}{1.7}   & \multicolumn{1}{c}{52.3~($\uparrow 3.7$)} & 77.7~($\uparrow 2.1$)  & \multicolumn{1}{c}{13.6} & \multicolumn{1}{c}{63.8~($\uparrow 0.0$)}  & 79.4~($\uparrow 2.0$) \\
\multicolumn{1}{c|}{What}  & \multicolumn{1}{c}{71.7}  & \multicolumn{1}{c}{59.4~($\uparrow 0.7$)} & 73.8~($\uparrow 0.4$)  & \multicolumn{1}{c}{18.5} & \multicolumn{1}{c}{60.8~($\uparrow 4.5$)}  & 81.0~($\uparrow 3.8$) \\ 
\multicolumn{1}{c|}{Where} & \multicolumn{1}{c}{5.7}   & \multicolumn{1}{c}{61.1~($\uparrow 3.8$)} & 72.8~($\uparrow 4.6$)  & \multicolumn{1}{c}{18.2} & \multicolumn{1}{c}{72.5~($\uparrow 2.3$)}  & 83.7~($\uparrow 0.8$) \\ 
\multicolumn{1}{c|}{Why}   & \multicolumn{1}{c}{4.3}   & \multicolumn{1}{c}{50.4~($\uparrow 0.0$)} & 61.0~($\downarrow 1.1$) & \multicolumn{1}{c}{15.7} & \multicolumn{1}{c}{41.5~($\downarrow 0.5$)} & 67.7~($\uparrow 1.5$) \\ 
\multicolumn{1}{c|}{How}   & \multicolumn{1}{c}{9.9}   & \multicolumn{1}{c}{52.6~($\uparrow 8.7$)} & 65.7~($\uparrow 7.5$)  & \multicolumn{1}{c}{15.3} & \multicolumn{1}{c}{47.3~($\uparrow 7.1$)}  & 68.3~($\uparrow 5.2$) \\ \bottomrule[1pt]
\end{tabular}
\vspace{-2mm}
\end{table}

\subsection{Implementation Details}
We implement our CADA based on Transformers Library \cite{li2020transformers}. We employ AdamW \cite{loshchilov2017decoupled} as our optimizer. For Molweni, we set the batch size as 16 and the learning rate as 5e-5 when BERT is used as the embedding layer. When we employ the ELECTRA backbone, the batch size and learning rate are set to 1e-5 and 12, respectively. For FriendsQA, we set the batch size as 8. The learning rate is set to 6e-6 and 4e-6 for the BERT-based CADA and the ELECTRA-based version respectively. In addition, the number of interlocutor property layers $L_1$ and discourse dependency layers $L_2$ ranges from 2 to 4 according to different PLMs. We train our models for 2 or 3 epochs. We run every CADA on 32G NVIDIA V100 GPUs. It takes about 1 hour to train CADA for one epoch on Linux.

\subsection{Main Results}
Table \ref{tab1} shows the test results obtained on both Molweni and FriendsQA. It can be observed that our CADA produces substantial improvements, compared to both BERT$_{large}$ and ELECTRA$_{large}$ baselines. More importantly, CADA achieves state-of-the-art performance for $F$1-score and EM in most cases, compared to the existing MDRC models. 

Specifically, we carry out the comparison with the neural models ESA \cite{ma2021enhanced} and SKIDB \cite{li2021self}. ESA is a strong graph-based neural model which leverages both the self-connected interlocutor graph and the mutual discourse dependency graph. Graph Convolutional Network (GCN) \cite{schlichtkrull2018modeling} is employed to strengthen the encoder for weighting available clues attentively. Particularly, ESA applies the masked attention modeling method conditioned on the structural information of the interlocutor graph. By contrast, SKIDB \cite{li2021self} employs a multi-task model which not only conducts the self-supervised interlocutor prediction but a key-utterance prediction. The shareable encoder for both tasks contributes to the joint incorporation of interlocutor property and key utterance information.

Our CADA is different from ESA and SKIDB. On the one hand, CADA straightforwardly refines the attention computation in the transformers conditioned on 1) utterance subgraphs of different individual interlocutors and 2) discourse dependency graphs, where GCN is not used. On the other hand, CADA additionally updates the PLM embedding layer using coreference resolution. According to the results in Table \ref{tab1}, ESA and SKIDB perform better than the baselines in most cases, except that SKIDB performs worse than the baselines on FriendsQA. On the contrary, they fail to outperform our CADA except that EM is considered for evaluation on FriendQA. Therefore, we suggest that, within the task-specific modeling method that is deployed upon the PLM layer, CADA has a relatively obvious superiority. Briefly, CADA outperforms ESA due to the fine-grained structural information integration, while it performs better than SKIDB because of the additional utilization of cross-utterance dependency features. More importantly, the awareness of coreference relationships in CADA enhances token-level encoding fundamentally.

Table \ref{tab22} provides a direct insight into the performance of CADA for different question types, where ELECTRA$_{large}$ is used as the baseline. It can be found that CADA stably outperforms the baseline for most question types except ``{\em Why}''-type questions. It proves that the perception of heterogeneous features (coreference features, internal features of an individual interlocutor, or interaction features among interlocutors) is beneficial for extractive MDRC. However, it is ineffective for answering ``{\em Why}''-type questions, which are more likely treated well using generative models (instead of extractive ones).

\subsection{Ablation Study}
We conduct an ablation analysis on Molweni and FriendsQA to verify the effects of each module of CADA. It is conducted by verifying performance variation when a module is removed, where the hyperparameters remain unchanged. Table \ref{tab2} shows the results of ablation experiments. It can be observed that the three modules have different domain-specific positive effects. The coreference-aware embedding layer is more effective than interlocutor property and discourse dependency modules when FriendsQA is considered. The condition is thoroughly reversed when Molweni is considered. It is most probably because that most of the dialogues in FriendsQA are colloquial conversations and are full of a larger number of pronouns.

\begin{table}[t]
\centering
\caption{\small{Ablation study on both corpora. IPM, DDM, and CAE are abbreviations of Interlocutor Property, Discourse Dependency modules, and Coreference-Aware Embedding layer, respectively.}}\label{tab2}
\tabcolsep=0.1cm
\begin{tabular}{lllll}
\toprule[1pt]
\multirow{2}{*}{Model} & \multicolumn{2}{c}{Molweni}  & \multicolumn{2}{c}{FriendsQA}  \\ & \multicolumn{1}{c}{EM}   & \multicolumn{1}{c}{F1}  & \multicolumn{1}{c}{EM}    & \multicolumn{1}{c}{F1}    \\ \hline
CADA (ELECTRA)             & \multicolumn{1}{c}{\textbf{59.8}} & \multicolumn{1}{c}{\textbf{73.6}} & 
\multicolumn{1}{c}{\textbf{59.2}}          & \multicolumn{1}{c}{\textbf{76.7}} \\ 
\quad w/o IPM                & \multicolumn{1}{c}{58.5}          & \multicolumn{1}{c}{72.6}          & 
\multicolumn{1}{c}{58.9} & \multicolumn{1}{c}{75.8}      \\ 
\quad w/o DDM                & \multicolumn{1}{c}{58.8}          & \multicolumn{1}{c}{72.5}          & 
\multicolumn{1}{c}{58.2}          & \multicolumn{1}{c}{75.5}        \\ 
\quad w/o CAE              & \multicolumn{1}{c}{59.2}          & \multicolumn{1}{c}{73.0}          & 
\multicolumn{1}{c}{58.2}          & \multicolumn{1}{c}{75.4}       \\ \bottomrule[1pt]
\end{tabular}
\vspace{-2mm}
\end{table}

\subsection{Effectiveness of CADA on Long Dialogues}
A long dialogue usually comprises a mass of utterances issued by a larger number of interlocutors. It possesses more complicated self-coherent and mutually-interactive relationships. 
We provide a study on the effects of MDRC models on different lengths of dialogues by using ELECTRA-based CADA. 
To be specific, we divide a test set into different subsets according to the number of utterances or interlocutors in dialogues and verify the performance of CADA on each subset. Molweni is considered in the experiments instead of FriendQA. It is because there are 40\% of dialogues in FriendQA contain overly-long contexts. External sliding windows are necessarily utilized for the separate encoding over them, which fundamentally obstructs the verification of the diversity of utterances and interlocutors.  

Table \ref{tab4} shows the performance variation over the subsets. It can be found that the performance of CADA degrades significantly when a dialogue involves more than 10 utterances or there are more than 3 interlocutors joining the multi-party dialogue. More seriously, the performance is reduced more severely when the Interlocutor Property Module (i.e., IPM for utterance profiling) is removed from CADA. Therefore, we suggest dealing with the long dialogues for MDRC is still challenging and should be paid more attention to in a real application scenario.

\begin{table}[t]
% \small
\setlength{\abovecaptionskip}{4pt}
\centering
\tabcolsep=0.1cm
\caption{\small{Performance of CADA (ELECTRA) on Molweni under different numbers of utterances or interlocutors.}}\label{tab4}
\begin{tabular}{llll}
\toprule[1pt]
\multicolumn{1}{c}{Molweni} & \multicolumn{1}{c}{EM}  & \multicolumn{1}{c}{F1}      \\ \hline 
\multicolumn{1}{l}{Number of utterances (CADA)}     & \multicolumn{1}{c}{}   & \multicolumn{1}{c}{} \\
$\le$7     & \multicolumn{1}{c}{61.0} & \multicolumn{1}{c}{75.9} \\  8$\sim$9  & \multicolumn{1}{c}{59.9} & \multicolumn{1}{c}{73.5} \\    $\ge$10      & \multicolumn{1}{c}{57.2} & \multicolumn{1}{c}{70.0} \\ \hline  
\multicolumn{2}{l}{Number of interlocutors (CADA)}      &     \\ 
1$\sim$2     & \multicolumn{1}{c}{60.7} & \multicolumn{1}{c}{74.4} \\ 
$\ge$3      & \multicolumn{1}{c}{59.1} & \multicolumn{1}{c}{72.9}\\ 
\multicolumn{2}{l}{Number of interlocutors (CADA w/o IPM)}     &      \\ 
1$\sim$2     & \multicolumn{1}{c}{60.8} & \multicolumn{1}{c}{73.4} \\ 
$\ge$3      & \multicolumn{1}{c}{57.4} & \multicolumn{1}{c}{71.9}\\ 
\bottomrule[1pt]
\end{tabular}
\end{table}

\subsection{Case Study}
\begin{figure}[t]
    \setlength{\belowcaptionskip}{-0.3cm}
    \centering
    \includegraphics[width=0.42\textwidth]{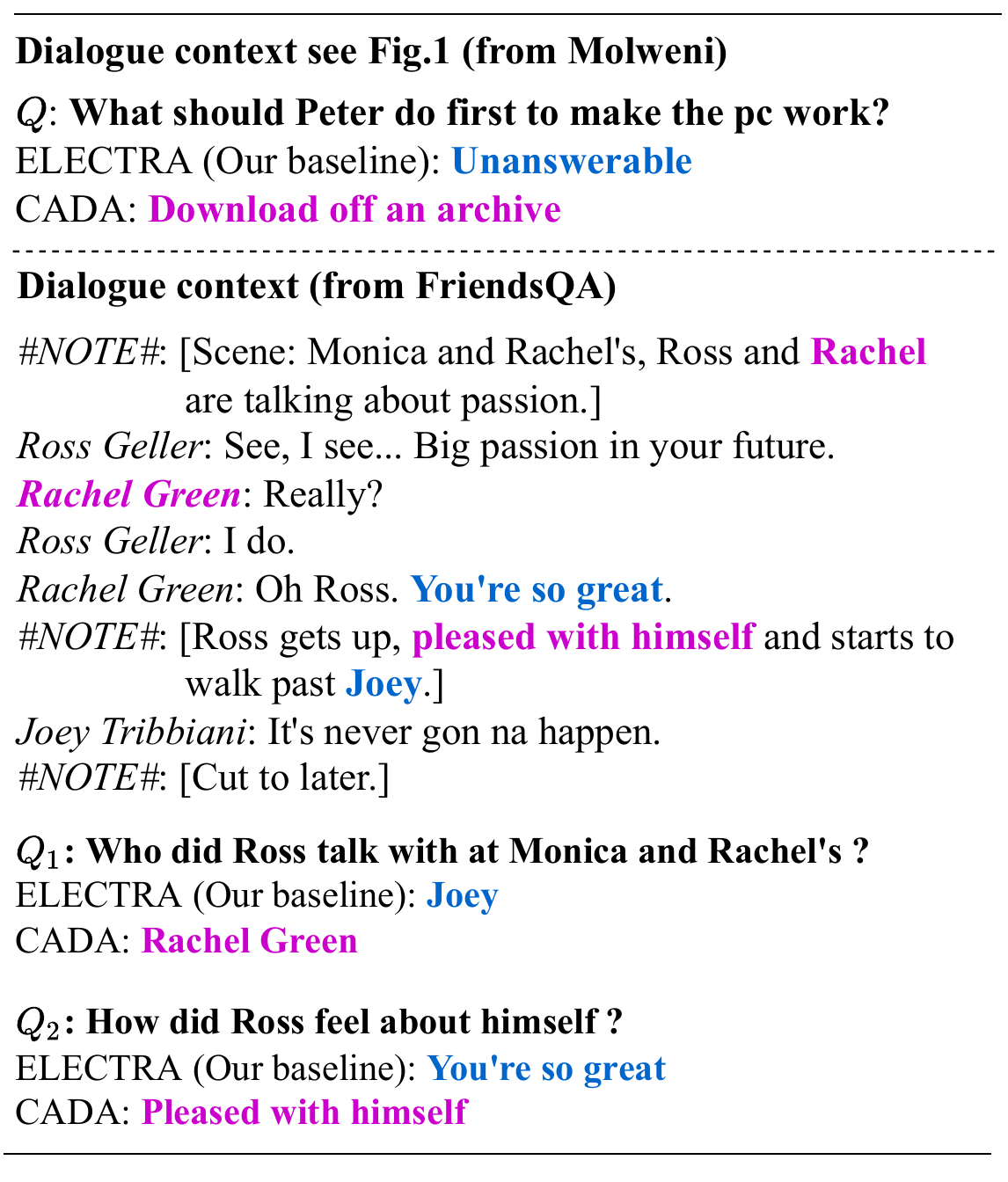}
    \caption{\small{Two selected cases from the Molweni and FriendsQA, where the correct answers are marked in purple and incorrect ones in blue.  \#\emph{NOTE}\# refers to the utterance that provides additional information (location, time, character activities, etc.).}}
    \label{figure4}
\end{figure}

We analyze a series of dialogues in Molweni and FriendsQA manually, so as to intuitively investigate the impacts of CADA compared to baselines. Fig.~\ref{figure4} shows two representative cases, for which CADA extracts precise answers while the ELECTRA baseline yields incorrect ones. Note that the dialogue context in the first case has been illustrated in Fig.~\ref{figure1}-({\tt b}), and thus it is omitted in Fig.~\ref{figure4} due to the page limitation. 

For the first case, the baseline predicts the question to be unanswerable, though CADA extracts the accurate answer. This is because CADA is able to better capture the discourse dependency from $U_5$ to $U_7$, as well as the co-referred clues (e.g., ``\emph{Peter}'' in $U_5$ and ``\emph{you}'' in $U_7$). This results from the coreference-aware encoder and discourse dependency module.

In the second case, for the question $Q_1$, the baseline predicts the answer to be ``\emph{Joey}''. It mistakenly implies that ``\emph{Ross}'' communicates with ``\emph{Joey}''. On the contrary, CADA precisely extracts the true answer ``\emph{Rachel Green}'' since it correctly captures interlocutor properties and interaction information. For the question ($Q_2$), the baseline gives the answer ``\emph{you're so great}''. It is the closely-related context to ``\emph{Ross}'', though it is uttered by \emph{Rachel Green} and has nothing to do with ``\emph{How Ross feels about himself}''. By contrast, CADA better captures the coreferential relation between \emph{Ross} and \emph{himself}. This clue supports the extraction of the correct answer.

\section{Conclusion}
We propose a novel MDRC model (namely CADA). CADA enables not only coreference-aware encoding but double-channel attention-weighted encoding. More importantly, it supports the separate implementation of utterance profiling (internal feature extraction) and interaction modeling (external feature extraction), as well as the combination of internal and external features. Experimental results on benchmark corpora demonstrate that CADA yields substantial improvements compared to the PLM-based baselines, and achieves state-of-the-art performance.  The codes are available at \url{https://github.com/YanLingLi-AI/CADA}.

However, CADA fails to maintain a stable performance on long multi-party dialogues. The sufferings comprise two main aspects, including the unawareness of complicated coherent relations within the long-term iteratively-occurred utterances of a single interlocutor, as well as that among a larger number of interlocutors. In the future, accordingly, we will study the modeling of memories, including the ones of self-coherent and interactively-coherent contents in historical utterances.

% \section*{Acknowledgments}
% The research is supported by the National Key R\&D Program of China (2020YFB1313601) and the National Science Foundation of China (62076174, 61836007).

\bibliographystyle{IEEEtran}
\bibliography{yingyong}

\end{document}